\documentclass[review]{elsarticle}

\usepackage{lineno,hyperref}
\usepackage{multirow}
\usepackage[table,xcdraw]{xcolor}
\usepackage{mathrsfs}
\usepackage{dutchcal}
\usepackage{amsmath}
\usepackage{diagbox}
\usepackage{graphicx}
\usepackage{caption}
\usepackage{float}

\journal{Journal of \LaTeX\ Templates}

\begin{document}

\captionsetup[figure]{labelfont={bf},labelformat={default},labelsep=period,name={Fig.}}
\begin{frontmatter}

\title{PAANet:Visual Perception based Four-stage Framework for Salient Object Detection using High-order Contrast Operator}

\author{Yanbo Yuan}
\author{Hua Zhong\corref{mycorrespondingauthor}}
\cortext[mycorrespondingauthor]{Corresponding author}
\ead{hzhong@mail.xidian.edu.cn}
\author{Haixiong Li}
\author{Xiao Cheng}
\author{Linmei Xia}
\address{School of Electronic Engineering, Xidian University}
\address{Xi’an, Shaanxi 710071, PR China}

\begin{abstract}
It is believed that human vision system (HVS) consists of pre-attentive process and attention process when performing salient object detection (SOD). Based on this fact, we propose a four-stage framework for SOD, in which the first two stages match the \textbf{P}re-\textbf{A}ttentive process consisting of general feature extraction (GFE) and feature preprocessing (FP), and the last two stages are corresponding to \textbf{A}ttention process containing saliency feature extraction (SFE) and the feature aggregation (FA), namely \textbf{PAANet}. According to the pre-attentive process, the GFE stage applies the fully-trained backbone and needs no further finetuning for different datasets. This modification can greatly increase the training speed. The FP stage plays the role of finetuning but works more efficiently because of its simpler structure and fewer parameters. Moreover, in SFE stage we design for saliency feature extraction a novel contrast operator, which works more semantically in contrast with the traditional convolution operator when extracting the interactive information between the foreground and its surroundings. Interestingly, this contrast operator can be cascaded to form a deeper structure and extract higher-order saliency more effective for complex scene. Comparative experiments with the state-of-the-art methods on 5 datasets demonstrate the effectiveness of our framework. 
\end{abstract}


\end{frontmatter}

\section{Introduction}

With the development of Artificial Intelligence (AI), Visual detection tasks have become more and more diversified, and have gradually become a basic problem in the field of computer vision. SOD needs to detect the most unique object or region and to obtain the valuable feature information in the scenario. Those information can be used in many image processing tasks such as image classification\cite{Murabito2018}, foreign object detection\cite{Wang2017c}, object tracking\cite{Wu2014}, image compression\cite{Nystrom2007}, image semantic segmentation\cite{Wei2017}, weak-supervised learning\cite{Lai2017}.

\begin{figure} [H]
	\centerline{\includegraphics[width=\columnwidth]{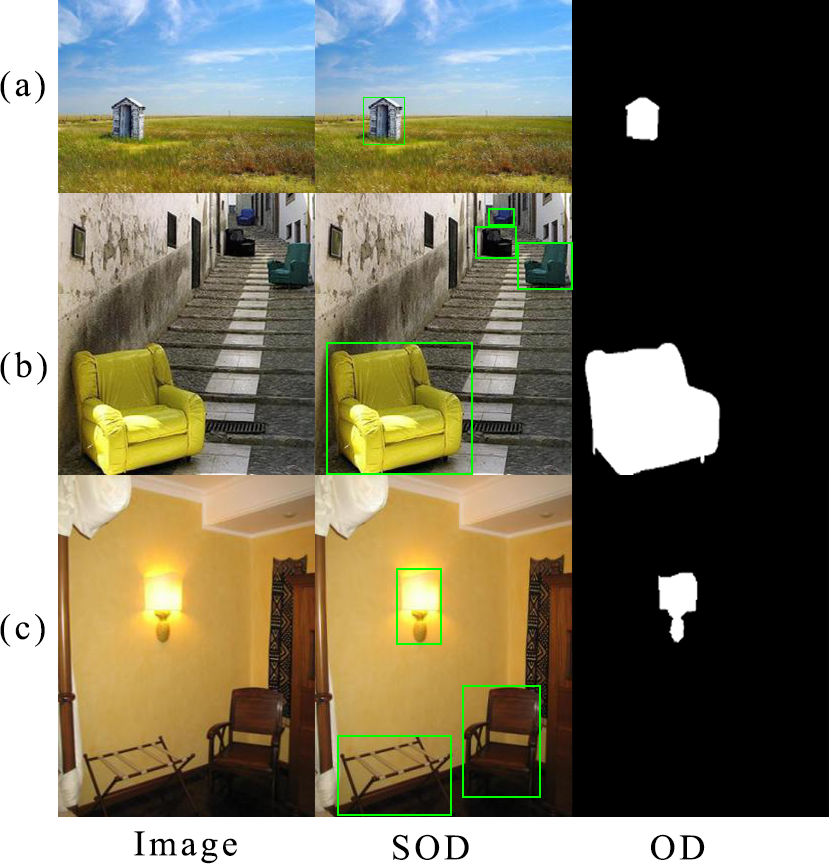}}
	\caption{Comparison of the results of SOD and OD.}
	\label{Fig1}
\end{figure}

Traditional object detection (OD) needs to find the location of all specific objects in the image. Instead, SOD needs to automatically detect the most obvious object in the image. As shown in Figure \ref{Fig1}, the first row is single object detection, and the results of the two methods are the same. The second row is multiple object detection of single category. OD detects all sofas, but SOD detects the most obvious yellow sofa. The last row is multiple object detection of different categories, OD detects all types of objects, while SOD detects the most prominent wall lamp.

Hou, etc.\cite{Hou2007} gives more detailed analysis of how the saliency detection process is achieved in human visual system. A visual processing involves two stages: First, the brain implements a simple, fast and parallel pre-attentive process of multiple tasks. At this time, the low-level basic features such as texture, edges, or colors obtained by the brain has been detected but has not yet been identified as a candidate object. The second stage is a continuous, slow and complex attention process. The brain will extract the special features we need. That is, feature processing and fusion are performed on the candidate features obtained in the first stage, so as to detect the object needed in the brain consciousness.

Based on ref.\cite{Hou2007}, we propose a four-stage salient object detection network. The saliency feature extraction module (SFEM) using high-order contrast operator can effectively reduce the impact of saliency uncertainty. At the same time, we propose a special network training method that can effectively save time in the training process.

Experiments prove that the method in this paper has excellent performance. Ultimately, main contributions of our work are summarized as follows:

\begin{enumerate}
	\itemsep=0pt
	\item According to the two processes of pre-attentive and attention, we design a salient object detection framework consisting of four stages: general feature extraction (GFE), feature preprocessing (FP), saliency feature extraction (SFE), and feature aggregation (FA).
	\item We design a saliency feature extraction module (SFEM) that uses contrast operator to learn the contrast information between candidate objects and background. The contrast operator can be cascaded to form a deeper structure and extract higher-order saliency which is meaningful for complex scene.
	\item We design a feature extraction structure consisting of GFE and FP. The backbone in our framework uses the pre-trained model and does not finetune, which improves the training speed while maintaining the completeness and versatility of the basic feature information.
	\item Experimental results proved the effectiveness of our framework on 5 saliency datasets. It not only outperforms state-of-the-art saliency methods in metrics, but also in visual comparison.
\end{enumerate}  

The rest of this paper is as follows. The related work is introduced in Chapter 2. In Chapter 3, we describe the framework structure and motivation with details. In Chapter 4, We make experiments to illustrate the effectiveness of PAANet. Finally, we summarize all the content in Chapter 5 and put forward prospects for future work.

\section{Related works}

\subsection{Backbone}

Most of the current SOD networks use bottom-up methods, which are a kind of passive, unconscious, open-loop, and data-driven methods. These methods train the classifier based on the image datasets and the training result of the classifier is determined by the characteristics and attributes of the training sets, such as MDF\cite{Li2016}, UCF\cite{Zhang2017a}, Amulet\cite{Zhang2017}, BASNet\cite{Qin2019}, Poolnet\cite{Liu2019}, MSWS\cite{Zeng2019a}, R3Net\cite{Deng2018}, U2Net\cite{Qin2020}, etc. They learn the common features of groundtruth, then find the regions which have common features, finally predict saliency maps.

These methods mostly use classification networks trained on ImageNet\cite{Deng2010} as backbone, then finetune it with saliency datasets to extract general features. MDF uses AlexNet\cite{Krizhevsky2017} as backbone. UCF, Amulet, NLDF\cite{Luo2017}, DSS+\cite{Hou2019}, RAS\cite{Chen2018}, BMPM\cite{Zhang2018c}, PicaNet\cite{Liu2018} and AFNet\cite{Feng2019} use VGGNet\cite{Simonyan2015} as backbone. BASNet, PoolNet, CPD\cite{Wu2019a}, DGRL\cite{Wang2018}, SRM\cite{Wang2017} and Capsal\cite{Zhang2019} use ResNet\cite{He2016}. MSWS uses Densenet-169\cite{Huang2017} and R3Net uses ResneXt101\cite{Xie2017}. But if the data distribution of the saliency dataset and ImageNet are quite different, training will be very inefficient. That was also mentioned in U2Net. U2Net designs a new RSU module as backbone trained from scratch to alleviate this problem. It improves the training effectiveness but increases training burden.

In our method, the design of the backbone corresponds to the pre-attentive process of HVS. Backbone is designed for generating general basic features that are suitable for most visual tasks. Therefore, we embed a well-trained backbone into the network, which holds powerful ability for extracting general basic features. Specially, our backbone doesn't participate in the fine-tuning of the whole salient object detection training. 

\subsection{SOD utilizing feature fusion}

In recent years, most existing methods use CNNs fusing multi-scale and multi-level features to improve the performance of saliency feature extraction due to its powerful ability for feature extraction. In which parameters of kernels need to be fully trained to be appropriate to the current task, such as U2Net, Capsal, DSS, CPD, UCF, etc. Specifically, Amulet integrates multi-level features into multiple resolutions to obtain different saliency prediction results, and finally merges them all to obtain the final saliency image. DVA\cite{Wang2018a} fuses deep features of different scales as saliency features after decoding by deconvolution layer. R3Net+ alternately fuses shallow and deep features to refine saliency features. In recent works, F3Net\cite{Wei2019} uses CFM to fuse features from different levels to mitigate the discrepancy between features. GCPANet\cite{Chen2020} designs a FIA module to integrate the low-level, high-level and global information in a mingle way, then uses a GCF module to generate context information to improve the completeness of saliency map. LDF\cite{Wei2020} encoders and fuses the output of backbone in four scales to generate body map, detail map and saliency map, then supervises them separately to improve the performance. 

In this paper, we introduce similarity comparison as prior knowledge and design a top-down SFEM utilizing semi-learning approach to extract saliency features on the basis of common features so as to highlight salient areas at each scale. Then, we process and aggregate those feature maps at different resolutions to predict the saliency map.

\section{Proposed Framework}

\subsection{Motivation}

\begin{figure}
	\centerline{\includegraphics[width=\columnwidth]{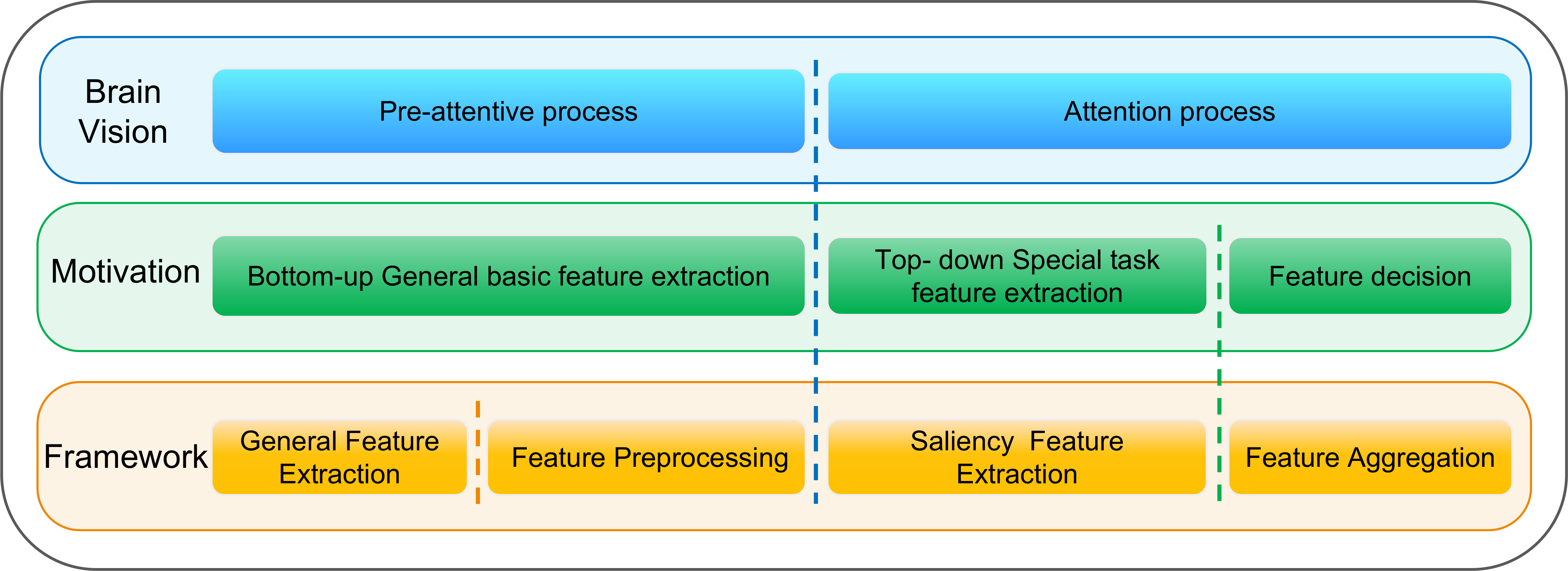}}
	\caption{The framework motivation of this paper. Blue is the process of human visual perception, green is the design motivation based on the process, and yellow is our framework structure.}
	\label{Fig2}
\end{figure}

The principle of our framework is shown in Figure \ref{Fig2}. It is mentioned in ref. \cite{Hou2007} that there are two processes in human visual perception named pre-attentive process and attention process. In our work, the pre-attentive process is implemented by bottom-up GFE, and the attention process is designed as top-down special task feature extraction and feature decision.

Bottom-up general extraction consists of GFE and FP. They are used to extract general visual features and modify them to suit for the saliency detection tasks. Top-down special task feature extraction is designed as the SFE that using high-level contrast operator. In training phase and test phase, the final FA is separately used to achieve multi-scale supervised learning and single-scale output.

\subsection{The components of PAANet}

\begin{figure} 
	\centerline{\includegraphics[width=\columnwidth]{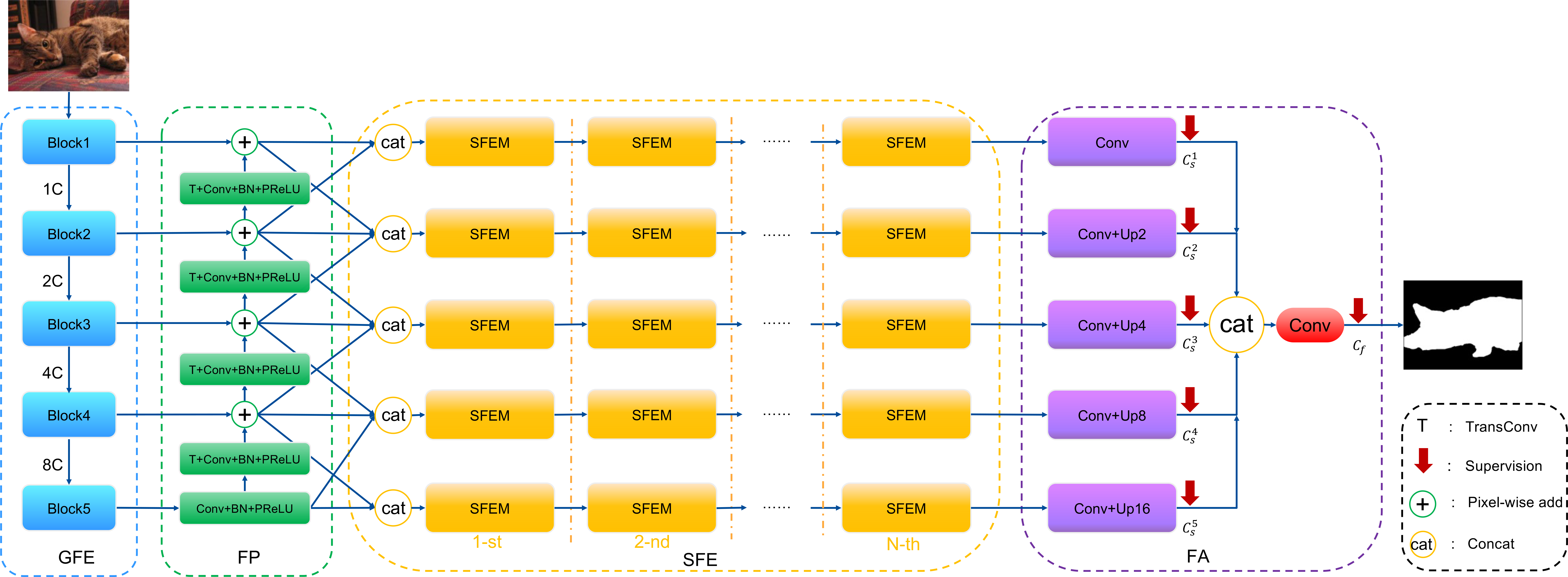}}
	\caption{The pipeline of our proposed four-stage framework for salient object detection using high-order contrast operator, which consists of four stages: GFE (blue), FP (green), SFE(yellow)and FA (purple). For the first stage, the factor C is the number of channels in the output feature maps of each block.}
	\label{Fig3}
\end{figure}

The framework we proposed is shown in Figure \ref{Fig3}. It consists of four stages— GFE, FP, SFE, FA. Features are extracted and processed on multiple scales in each stage, and finally aggregated into a saliency map with the same scale as the input image.

\subsubsection{General Feature Extraction (GFE)}

GFE, the backbone of our framework, extracts all general features in an image, corresponding to the leftmost blue module in Figure \ref{Fig3}, which is a part of the visual pre-attentive process.

ImageNet almost contains all natural scenes. Thus, our backbone fully trained in this dataset will obtain complete and abundant detailed basic features, and has excellent general feature extraction capabilities in natural images. Saliency datasets we use in our work are also natural scenes, so the backbone needs not to be finetuned to maintain its ability.

We use Resnext-101 as the backbone in our framework. Five-scale blocks are used to extract multi-scale features from the image, which will be refined as the candidate features for subsequent work.

\begin{equation}
	F_\text{gen}=\Psi(x)
\end{equation}

Where $x$ is the input image, $\Psi(x)$ is the reflection of backbone to extract general feature information of x.

\subsubsection{Feature Preprocessing (FP)}

FP is a bridge between the GFE and SFE. In this stage, the general features extracted from backbone are preliminarily modified to be more suitable for the SOD task. The module (green) shown in Figure \ref{Fig3}, as a part of the visual pre-attentive process, is also based on bottom-up method of visual feature extraction. 

\begin{figure} 
	\centerline{\includegraphics[width=\columnwidth]{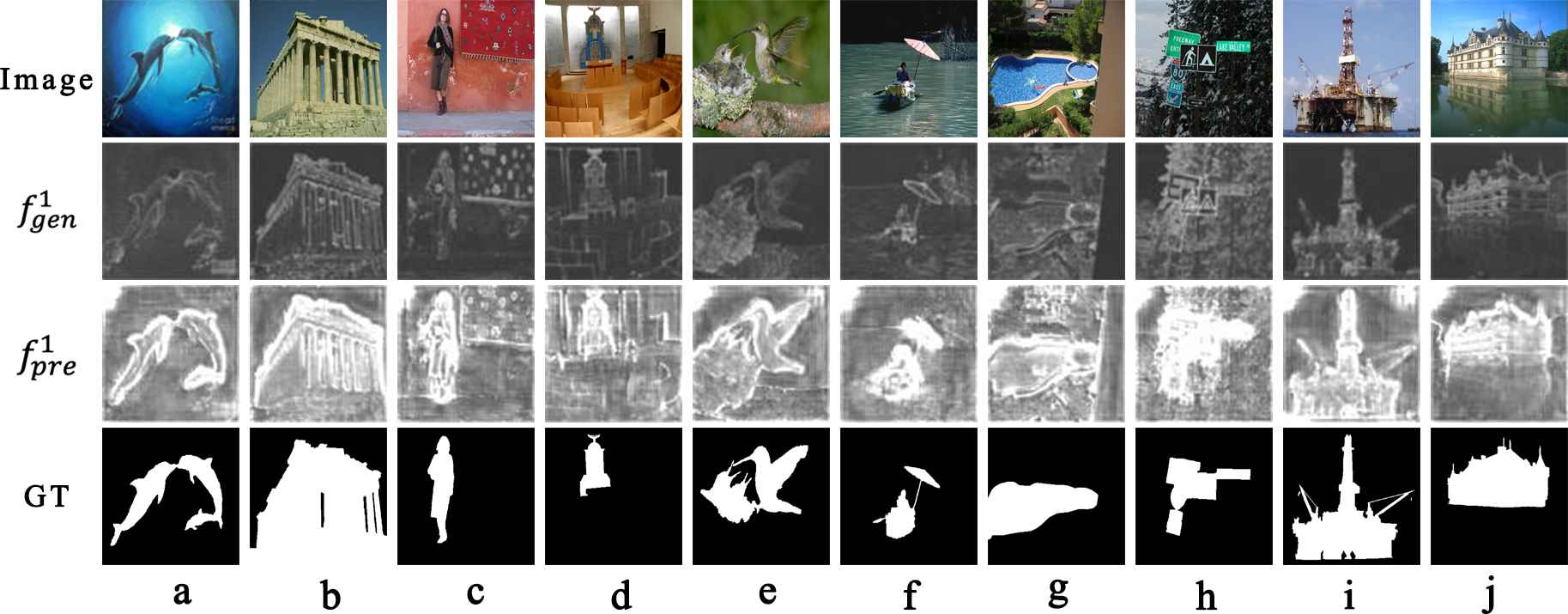}}
	\caption{Comparison between the feature maps of GFE and FP. $f_{gen}^1$ is the first branch of GFE outputs, $f_{pre}^1$ is the first branch of FP outputs.}
	\label{Fig4}
\end{figure}

As shown in the Figure \ref{Fig4}, this stage processes features at five different scales. The connection between blue and green phases is similar to the jump connection in UNet\cite{Ronneberger2015}. However, we use pixel-wise add instead of concatenation to connect the feature maps. Details are formulated as follows:

\begin{equation}
	f^i_\text{pre}=
	\begin{cases}
		M^i_\text{FP}(f^{i+1}_\text{gen})+f^i_\text{gen}&{i=1,2,3}\\
		M^{i-1}_\text{FP}(M^i_\text{FP}(f^{i}_\text{gen}))+f^{i-1}_\text{gen}&{i=4}\\
		M^i_\text{FP}(f^{i}_\text{gen})&{i=5}
	\end{cases}
\end{equation}

Where $f^i_{gen} (i=1, 2, …, 5)$ is $i_\text{th}$ scale in $F_{gen}$, $ M^i_{FP} $ is $i_\text{th}$ module in FP. $M^i_{FP}$ consists of a transposed convolution layer, 2 convolution layers, a PReLU\cite{He2015} layer and a Batch Normalization\cite{Ioffe2015} layer to generate preprocessed feature maps of 1, 1/2, 1/4, 1/8, 1/16 resolutions respectively. Specially, when $ i=5 $, transposed convolution is discarded. Then the preprocessed features will be added with corresponding $f^i_{gen}$. Finally, the fusion results $f^i_{pre}$ are used as the input of higher resolution feature preprocessing.

We show a set of comparative figures to illustrate the effectiveness of FP. As we can see in Figure \ref{Fig4}, taking the first branch (with maximum resolution) as an example, FP strengthens the characteristics of the objects especially those with complex lines and weaken the surroundings, the focus regions become more compact compared with the former stage . In addition, compared with the outputs of GFE, the outputs of FP have higher intensity. In other words, according to FP, the feature maps will be revised to be more suitable for subsequent stages and closer to GT.

\subsubsection{Saliency Feature Extraction (SFE)}

\begin{figure}
	\centerline{\includegraphics[width=\columnwidth]{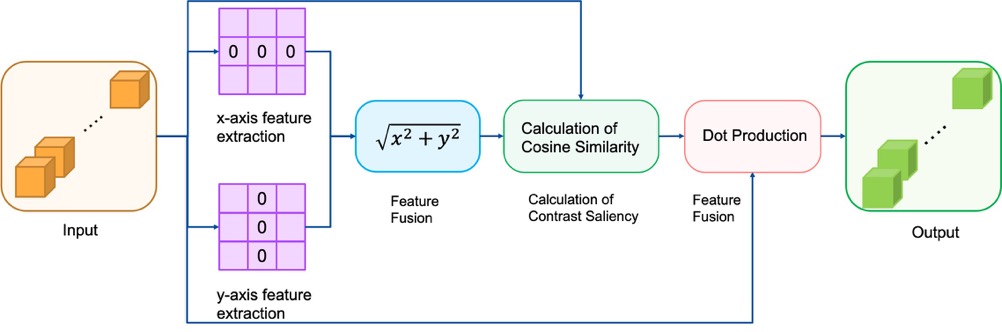}}
	\caption{Contrast operator extracts X-axis and Y-axis environmental feature by two semi-learning convolution kernels (purple), calculates the difference by cosine similarity and fuses it with the input to generate special features.}
	\label{Fig5}
\end{figure}

SFE stage (orange in Figure \ref{Fig3}) is a part of attention process in human vision perception. It bases on Top-down special feature extraction method. Our framework uses a high-order cascade approach. Each order contains 5 saliency feature extractor modules (SFEMs) to extract interactive feature information at different resolutions accomplished by corresponding contrast operators. These cascaded modules are the key to improve the performance of our framework. The internal structure of each contrast operator is shown in Figure \ref{Fig5}.

We evaluate the saliency of this object by the similarity between the salient candidate object and the surrounding environment. As the experiment shows, whether the object in an image is salient depends on the difference between the object and the environment. The greater the difference, the more salient the object, and vice versa. Different from the current multi-scale feature fusion method, we found that the interactive features between different regions in the image can represent the difference between the object and the background, so we use it to describe the saliency of the object.
 
Each order contains five SFEMs using contrast operators, which extract saliency features of different scales respectively. We process the output feature maps of previous FP stage to be the input of each SFE module. First, the feature maps are down-sampled or bi-linearly up-sampled to the same size. Then they are combined into a set of feature maps through concatenation operation. Finally, we input them to the corresponding SFE modules. This stage can be formulated as follow:

\begin{equation}
	f^i_\text{saliency}=
	\begin{cases}
		M^{(n)}_\text{SFE}(\text{cat}(f^{i}_\text{pre}, up(f^{i+1}_\text{pre})))&{i=1}\\
		M^{(n)}_\text{SFE}(\text{cat}(\text{down}(f^{i-1}_\text{pre}),f^{i}_\text{pre},\text{up}(f^{i+1}_\text{pre}))&{i=2,3,4}\\
		M^{(n)}_\text{SFE}(\text{cat}(\text{down}(f^{i-1}_\text{pre}), f^{i}_\text{pre}))&{i=5}
	\end{cases}
\end{equation}

Where $M^{(n)}_\text{SFE}$ is the module in SFE stage, in which n means the number of the orders. We choose n=3 in our framework. The larger the n, the deeper and more abundant features the stage can obtain. up and down are separately standing for 2x up-sampling or down-sampling.

 The details of this operator are described in Figure \ref{Fig5}. First of all, the input feature maps will pass through two parallel semi-learning 3*3 convolutional layers, which similar to the Sobel operator. Sobel operator is an isotropic pixel-wise 3x3 image gradient operator to compute an approximation of the gradient of the image intensity function. The semi-learning convolution kernel we designed is to extract the image feature vector around the central area in the X-axis direction and the Y-axis direction. Only the remaining 6 parameters are learned of each convolution kernel channel.
 
Then, we fuse features from X and Y axis by calculating the square sum of the corresponding elements to obtain feature maps containing environmental information. Subsequently, the cosine similarity is used for calculating the saliency of the target area and the surrounding environment to generate a weight map with the same size as the input feature map. Finally, the weight map used as MASK is fused with the input by multiplication to be the output saliency feature map.

\subsubsection{Feature Aggregation (FA)}

FA is used to aggregate saliency feature maps of every scale, and optimize the results of the previous module. In the training phase of the framework, five different scales of saliency results are up sampled to the same scale to calculate loss with GT.

\begin{equation}
	f_\text{fuse}=\text{conv}(\text{cat}_{i=1,2,...,5}(M_\text{FA}(f^i_\text{saliency})))
\end{equation}

$M_\text{FA}$ is utilized for the input feature of each scale to normalize channel and predict saliency map. As Figure \ref{Fig3} shows, Each module ($M_\text{FA}$) in this stage is composed of Conv and $Up_\text{j} (j={2,4,8,16}$ represents the degree of upsampling), where Conv consists of a convolution layer, a BN layer and a PReLU layer, and $Up_\text{j}$ means that the feature map obtained after the convolution operation is up-sampled to obtain the feature map of the original image resolution.

\subsection{Training and Supervision}

\subsubsection{Training strategy}

We designed a special training method for the framework proposed in this paper. The parameters in the backbone are fixed and do not update in the backward process during training. Because the backbone has learned abundant and versatile basic visual feature information in ImageNet. This design maintains the powerful capability to extract general features, which can also reduce the amount of training parameters to speed up the convergence of the framework. In Chapter 4.4, we conduct an ablation study to demonstrate the effectiveness of our training strategy.

\subsubsection{Loss}

In order to get more exquisite features, we use Deep Supervision\cite{Lee2015} in this Network. We supervise at six different scales in this Network, including the final fusion saliency feature map $f_\text{fuse}$ of the network and the five saliency feature maps ${f^{i}_\text{saliency},i=1,2,…,5}$ in the FA stage. We use Binary cross entropy (BCE) to calculate loss for each scale. Simply, we use $f={{f_\text{fuse}, f^{i}_\text{saliency} i=1,2,…,5}}$ to normalize the following formula. 

\begin{equation}
	l=-\sum\nolimits^{(H,W)}_{(i,j)}[G_{(i,j)}\text{log}(f_{(i,j)})+(1-G_{(i,j)})\text{log}(1-f_{(i,j)})]
\end{equation}

Where $(H,W)$ is the height and width of the input image. $G_{(i,j)}$  and $f_{(i,j)}$ represent the pixel values of Ground Truth and the predicted saliency probability map at coordinates $(i,j)$, respectively. 
In the overall training phase, our loss function is defined as:

\begin{equation}
	L=\sum\limits^{M}_{i=1}W_s^il_s^i+W_fl_f
\end{equation}

Where $L$ is the total loss of the network. $M$ is the number of scales used in the network which is set to 5. $i=[1,2,3,4,5]$ represents the index of the scale. $l^i_{s}$ represents the loss of the side output saliency map on scale i. $l_{f}$ represents the loss of the fusion saliency map. $W^{i}_{s}$ and $W_{f}$ are corresponding loss weights, which are all set to 1 in this network. During the test phase, we apply the fusion output $f_\text{fuse}$ as the final saliency map result.

\section{Experimental results}

\subsection{Datasets}

We train the proposed framework on DUTS-TR\cite{Wang2017a} dataset. DUTS-TR is currently the largest and most commonly used dataset for SOD, containing 10553 training images and respective ground truth.

We evaluate PAANet on five benchmark datasets, including DUT-OMRON\cite{Yang2013}, HKU-IS\cite{Li2015}, ECSSD\cite{Yan2013}, SOD\cite{Movahedi2010}, PASCAL-S\cite{Li2014}. DUT-OMRON contains 5168 natural images, most of which have one or two objects with complex structure. HKU-IS includes 4447 natural images, with multiple foreground objects in every image. ECSSD consists of 1000 natural images, most of which have large and structurally complex foreground object. SOD is a small-scale challenging dataset, which is originally designed for image segmentation and consists of 300 images. The images have low contrast and structure-complex objects. PASCALS-S is the most challenging dataset, including 850 images, most of which have complex foreground objects and cluttered background.

\subsection{Experimental settings}

\subsubsection{Implementation details}

During training and inference, the hardware environment mainly includes an Intel i5-6300K@3.72GHz CPU with 16GB of RAM, and an NVIDIA GTX 1070Ti with 8GB memory. To accelerate computation, we use CUDA and cudnn.

We adopt Pytorch framework to implement our proposed method. During training phase, the base learning rate, gamma, momentum, weight decay and epochs are set to 10-4, 0.1, 0.9, 0.0005, 10, respectively. 

\subsubsection{Evaluation metrics}

The output of such networks is a probability map with the same resolution as the input image. The value of each pixel is in the range of [0,1], representing as the probability that the pixel belongs to the saliency region. A threshold is generally set to identify whether a pixel belongs to object, the pixel greater than the threshold is positive which indicates an object pixel, otherwise, it belongs to background.

We evaluate our method using five numerical metrics and five curve metrics in experiments, while comparing with other models on these ten measurements, respectively maximal F-measure (MaxFm), mean F-measure(MeanFm), E-measure(EM)\cite{Fan2018}, Structure-measure(S-measure)\cite{Fan2017}, MAE, Precision-Recall (PR) curve, Receiver Operating Characteristic(ROC) curve, Em curve, Fm curve, Sm-MAE curve.

\subsection{Comparison with the State-of-the-arts}

We compare our method with 9 state-of-the-art methods on five datasets (In the first column of the Table \ref{Table1} for compared saliency algorithm), including Capsal, LDF, F3Net, DGRL, BASNet, R3Net, CPD, MSWS, GCPANET.

To ensure the fairness of comparison, the outputs of these approaches are transformed to the range of [0,255]. And we use the released code of  ref. \cite{Fan2018} to test five numerical metrics and five curve metrics.

\subsubsection{Quantitative comparisons}

Quantitative comparisons of the MAE, MaxFm, MeanFm, Em, Sm results based on five datasets is shown in Table \ref{Table1}.

Our method achieves the first for most of the metrics, and the second or third for most of the remaining metrics on the HKU-IS, SOD, PASCAL-S and DUT-OMRON datasets. We also achieve the second or third for most of the metrics on ECSSD.

Compared with these existing state-of-the-art methods, we have the greatest improvement of the metrics on PASCAL-S datasets, which contains multiple salient objects and complex background. On this dataset, we achieve the first for five metrics, including MAE, MaxFm, MeanFm, MaxE and MeanEm, which proves our method performs well for complex background and multiple salient object detection.

\begin{table*}[]
	\caption{Comparison of our method and 9 SOTA methods. MAE (↓), MaxFm  (↑),  MeanFm (↑), MaxEm (↑), MeanEm (↑), SM (↑). The best three results are highlighted in red, green and blue. - : with no data.}
\centering
\resizebox{\textwidth}{!}
	{
	\begin{tabular}{cccccccccccc}
		\hline
		Datasets                    & Criteria & $\text{DGRL}^{\text{CVPR18}}$                                 & $\text{R3Net}^{\text{IJCAI18}}$                                & $\text{BASNet}^{\text{CVPR19}}$                                & $\text{Capsal}^\text{{CVPR19}}$ & 
		$\text{CPD}^{\text{CVPR19}}$  & $\text{MSWSNet}^{\text{CVPR19}}$                               & $\text{F3Net}^{\text{AAAI20}}$                                & $\text{GCPANet}^{\text{AAAI20}}$                               & $\text{LDFNet}^{\text{CVPR20}}$                               & Ours                                   \\ \hline
		& MAE      & 0.0632                                & 0.0625                                 & 0.0565                                 & 0.1046   & 0.056  & 0.1077                                 & {\color[HTML]{00B050} \textbf{0.0526}} & 0.0566                                 & {\color[HTML]{C00000} \textbf{0.0517}} & {\color[HTML]{0070C0} \textbf{0.0541}} \\ \cline{2-12} 
		& MaxFm    & 0.7406                                & 0.7604                                 & {\color[HTML]{00B050} \textbf{0.7792}} & 0.5218   & 0.7536 & 0.6765                                 & {\color[HTML]{0070C0} \textbf{0.778}}  & 0.775                                  & {\color[HTML]{C00000} \textbf{0.7818}} & 0.7677                                 \\ \cline{2-12} 
		& MeanFm   & 0.727                                 & 0.7472                                 & {\color[HTML]{0070C0} \textbf{0.767}}  & 0.5061   & 0.7421 & 0.5976                                 & 0.7662                                 & 0.7558                                 & {\color[HTML]{C00000} \textbf{0.77}}   & {\color[HTML]{00B050} \textbf{0.7673}} \\ \cline{2-12} 
		& MaxEm    & 0.8534                                & 0.8573                                 & {\color[HTML]{0070C0} \textbf{0.8715}} & 0.6783   & 0.8683 & 0.8164                                 & {\color[HTML]{00B050} \textbf{0.8716}} & 0.8685                                 & 0.8695                                 & {\color[HTML]{C00000} \textbf{0.8722}} \\ \cline{2-12} 
		& MeanEm   & 0.8449                                & 0.8528                                 & {\color[HTML]{00B050} \textbf{0.865}}  & 0.6668   & 0.8468 & 0.7288                                 & 0.8647                                 & 0.8536                                 & {\color[HTML]{00B050} \textbf{0.865}}  & {\color[HTML]{C00000} \textbf{0.8677}} \\ \cline{2-12} 
		\multirow{-6}{*}{DUT-OMRON} & Sm       & 0.8097                                & 0.8166                                 & 0.8362                                 & 0.6831   & 0.8248 & 0.7559                                 & {\color[HTML]{0070C0} \textbf{0.8385}} & {\color[HTML]{00B050} \textbf{0.8389}} & {\color[HTML]{C00000} \textbf{0.8392}} & 0.8206                                 \\ \hline
		& MAE      & 0.0426                                & 0.0402                                 & 0.037                                  & 0.085    & 0.0371 & 0.0986                                 & {\color[HTML]{C00000} \textbf{0.0332}} & 0.0356                                 & {\color[HTML]{00B050} \textbf{0.0335}} & {\color[HTML]{0070C0} \textbf{0.0346}} \\ \cline{2-12} 
		& MaxFm    & 0.9158                                & 0.9257                                 & 0.9312                                 & 0.8033   & 0.9261 & 0.8588                                 & 0.9353                                 & {\color[HTML]{0070C0} \textbf{0.9362}} & {\color[HTML]{C00000} \textbf{0.9377}} & {\color[HTML]{00B050} \textbf{0.9364}} \\ \cline{2-12} 
		& MeanFm   & 0.9005                                & 0.9168                                 & 0.9169                                 & 0.777    & 0.9131 & 0.7625                                 & {\color[HTML]{00B050} \textbf{0.9249}} & 0.9159                                 & {\color[HTML]{C00000} \textbf{0.9273}} & {\color[HTML]{0070C0} \textbf{0.9198}} \\ \cline{2-12} 
		& MaxEm    & 0.9462                                & 0.9487                                 & 0.951                                  & 0.854    & 0.9511 & 0.9093                                 & {\color[HTML]{00B050} \textbf{0.9547}} & {\color[HTML]{C00000} \textbf{0.9553}} & {\color[HTML]{0070C0} \textbf{0.9538}} & 0.9494                                 \\ \cline{2-12} 
		& MeanEm   & 0.9383                                & 0.9437                                 & 0.9432                                 & 0.8362   & 0.942  & 0.7905                                 & {\color[HTML]{00B050} \textbf{0.9479}} & 0.9447                                 & {\color[HTML]{C00000} \textbf{0.9482}} & {\color[HTML]{0070C0} \textbf{0.9452}} \\ \cline{2-12} 
		\multirow{-6}{*}{ECSSD}     & Sm       & 0.906                                 & 0.9102                                 & 0.9163                                 & 0.8214   & 0.9182 & 0.8276                                 & 0.9242                                 & {\color[HTML]{C00000} \textbf{0.9268}} & {\color[HTML]{0070C0} \textbf{0.9245}} & {\color[HTML]{00B050} \textbf{0.9249}} \\ \hline
		& MAE      & 0.115                                 & 0.1414                                 & 0.1217                                 & 0.1441   & 0.1202 & {\color[HTML]{C00000} \textbf{0.1804}} & {\color[HTML]{0070C0} \textbf{0.1109}} & {\color[HTML]{00B050} \textbf{0.1083}} & 0.1118                                 & 0.1069                                 \\ \cline{2-12} 
		& MaxFm    & {\color[HTML]{0070C0} \textbf{0.839}} & 0.8078                                 & 0.8306                                 & 0.7832   & 0.8226 & 0.7849                                 & 0.8384                                 & {\color[HTML]{00B050} \textbf{0.8441}} & 0.8388                                 & {\color[HTML]{C00000} \textbf{0.8473}} \\ \cline{2-12} 
		& MeanFm   & 0.8293                                & 0.8038                                 & 0.8213                                 & 0.7621   & 0.8152 & 0.6857                                 & 0.8336                                 & {\color[HTML]{0070C0} \textbf{0.8339}} & {\color[HTML]{00B050} \textbf{0.8343}} & {\color[HTML]{C00000} \textbf{0.843}}  \\ \cline{2-12} 
		& MaxEm    & 0.8473                                & 0.8073                                 & 0.8344                                 & 0.8011   & 0.8388 & 0.8177                                 & {\color[HTML]{0070C0} \textbf{0.8535}} & {\color[HTML]{00B050} \textbf{0.8549}} & 0.8498                                 & {\color[HTML]{C00000} \textbf{0.8605}} \\ \cline{2-12} 
		& MeanEm   & 0.8354                                & 0.7887                                 & 0.8214                                 & 0.7811   & 0.8197 & 0.6937                                 & {\color[HTML]{0070C0} \textbf{0.8355}} & {\color[HTML]{00B050} \textbf{0.8396}} & 0.834                                  & {\color[HTML]{C00000} \textbf{0.8427}} \\ \cline{2-12} 
		\multirow{-6}{*}{PASCAL-S}  & Sm       & 0.7959                                & 0.7547                                 & 0.7846                                 & 0.7588   & 0.7895 & 0.7274                                 & {\color[HTML]{00B050} \textbf{0.802}}  & {\color[HTML]{C00000} \textbf{0.8094}} & 0.7984                                 & {\color[HTML]{0070C0} \textbf{0.8}}    \\ \hline
		& MAE      & 0.0374                                & 0.0338                                 & 0.0322                                 & 0.0607   & 0.0342 & 0.0858                                 & {\color[HTML]{0070C0} \textbf{0.028}}  & 0.0316                                 & {\color[HTML]{FF0000} \textbf{0.0275}} & {\color[HTML]{C00000} \textbf{0.0275}} \\ \cline{2-12} 
		& MaxFm    & 0.9023                                & 0.9102                                 & 0.9191                                 & 0.8406   & 0.9108 & 0.8352                                 & 0.9254                                 & {\color[HTML]{0070C0} \textbf{0.9268}} & {\color[HTML]{C00000} \textbf{0.9288}} & {\color[HTML]{00B050} \textbf{0.9283}} \\ \cline{2-12} 
		& MeanFm   & 0.8851                                & 0.8996                                 & 0.9026                                 & 0.8116   & 0.8926 & 0.7344                                 & {\color[HTML]{0070C0} \textbf{0.9099}} & 0.9013                                 & {\color[HTML]{00B050} \textbf{0.9134}} & {\color[HTML]{C00000} \textbf{0.92}}   \\ \cline{2-12} 
		& MaxEm    & 0.9488                                & 0.9479                                 & 0.9515                                 & 0.8944   & 0.9503 & 0.9076                                 & 0.9576                                 & {\color[HTML]{00B050} \textbf{0.958}}  & {\color[HTML]{0070C0} \textbf{0.9579}} & {\color[HTML]{C00000} \textbf{0.9599}} \\ \cline{2-12} 
		& MeanEm   & 0.9392                                & 0.9425                                 & 0.9432                                 & 0.8762   & 0.9386 & 0.7867                                 & {\color[HTML]{0070C0} \textbf{0.9518}} & 0.9448                                 & {\color[HTML]{00B050} \textbf{0.9529}} & {\color[HTML]{C00000} \textbf{0.9533}} \\ \cline{2-12} 
		\multirow{-6}{*}{HKU-IS}    & Sm       & 0.8968                                & 0.8998                                 & 0.909                                  & 0.85     & 0.9056 & 0.8183                                 & {\color[HTML]{0070C0} \textbf{0.9172}} & {\color[HTML]{C00000} \textbf{0.9205}} & {\color[HTML]{00B050} \textbf{0.9196}} & {\color[HTML]{0070C0} \textbf{0.9172}} \\ \hline
		& MAE      & 0.1059                                & 0.1223                                 & 0.1136                                 & 0.1517   & 0.1116 & 0.1699                                 & {\color[HTML]{0070C0} \textbf{0.0994}} & {\color[HTML]{C00000} \textbf{0.0902}} & -                                      & {\color[HTML]{00B050} \textbf{0.0953}} \\ \cline{2-12} 
		& MaxFm    & 0.8048                                & 0.8211                                 & 0.8052                                 & 0.6529   & 0.8139 & 0.7715                                 & {\color[HTML]{0070C0} \textbf{0.8262}} & {\color[HTML]{00B050} \textbf{0.8272}} & -                                      & {\color[HTML]{C00000} \textbf{0.8372}} \\ \cline{2-12} 
		& MeanFm   & 0.7905                                & {\color[HTML]{00B050} \textbf{0.8096}} & 0.79                                   & 0.6356   & 0.7702 & 0.6311                                 & 0.8059                                 & {\color[HTML]{0070C0} \textbf{0.8083}} & -                                      & {\color[HTML]{C00000} \textbf{0.8195}} \\ \cline{2-12} 
		& MaxEm    & 0.837                                 & 0.8348                                 & 0.829                                  & 0.7227   & 0.8484 & 0.8208                                 & {\color[HTML]{0070C0} \textbf{0.8554}} & {\color[HTML]{C00000} \textbf{0.8618}} & -                                      & {\color[HTML]{00B050} \textbf{0.8593}} \\ \cline{2-12} 
		& MeanEm   & 0.8065                                & 0.804                                  & 0.7979                                 & 0.6957   & 0.7776 & 0.6561                                 & {\color[HTML]{0070C0} \textbf{0.8143}} & {\color[HTML]{00B050} \textbf{0.8235}} & -                                      & {\color[HTML]{C00000} \textbf{0.8237}} \\ \cline{2-12} 
		\multirow{-6}{*}{SOD}       & Sm       & 0.7731                                & 0.7624                                 & 0.7694                                 & 0.6982   & 0.7672 & 0.7                                    & {\color[HTML]{00B050} \textbf{0.7896}} & {\color[HTML]{C00000} \textbf{0.8043}} & -                                      & {\color[HTML]{0070C0} \textbf{0.7845}} \\ \hline
	\end{tabular}
}
	\label{Table1}
\end{table*}

\subsubsection{Qualitative comparisons}

\begin{figure*}
	\centerline{\includegraphics[width=0.98\columnwidth]{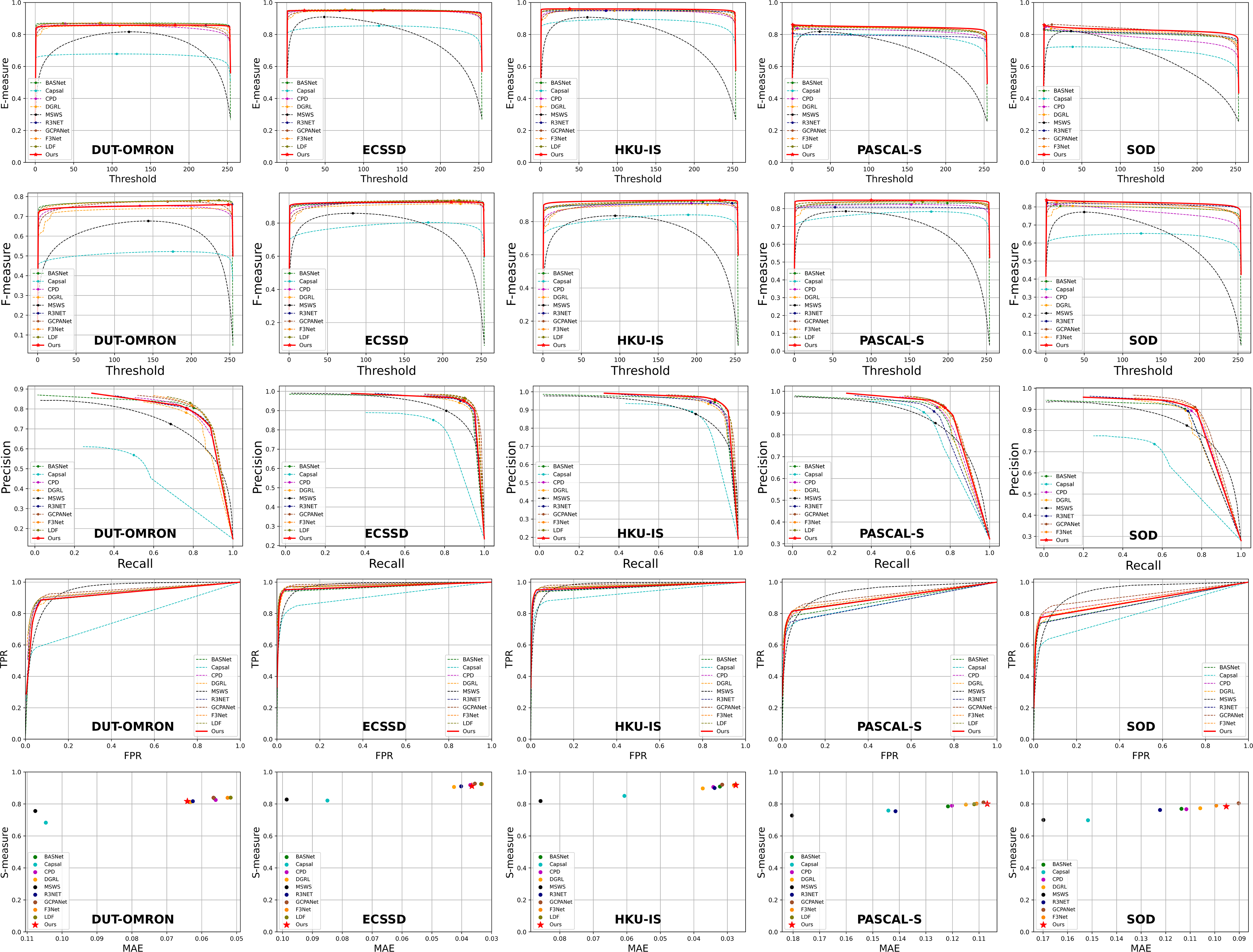}}
	\caption{Comparisons of EM curves, PR curves, F-measure curves, ROC Curves, S-measure curves on five common saliency datasets.}
	\label{Fig6}
\end{figure*}

We show the PR-curves, Em Curves, Fm Curves, ROC Curves and Sm-MAE curves contained all the methods on five datasets in Figure \ref{Fig6}. The red line represents our method. In the EM curves and FM curves of the first and second rows, our approach shows great performance for all various thresholds on all the test datasets. In the PR curves of the third row, our approach achieves the first or second results on HKU-IS, SOD, PASCAL-S and ECSSD with high recall and precision. The performance of the PR curves of ours is decent on DUT-OMRON, and keep high recall. It indicates that our model predicts delicate saliency maps. Compared with other methods, such as Capsal, LDF and F3Net, our approach shows a higher precision at the same recall value. In other words, the salient object detection results of our model have less errors. The comparisons in terms of ROC curves are given in the fourth row. the FPR value of our model is low on ECSSD and HKU-IS, which means less detection errors. Our approach reaches top3 in ROC performance. In the Sm-MAE figure of the fifth row, our method achieves the first on HKU-IS and PASCAL-S and the second on SOD and ECSSD. It shows that our model gets low error and meanwhile handles the problem of structural similarities well.

\subsubsection{Vision comparison}

\begin{figure*}
	\centerline{\includegraphics[width=0.98\columnwidth]{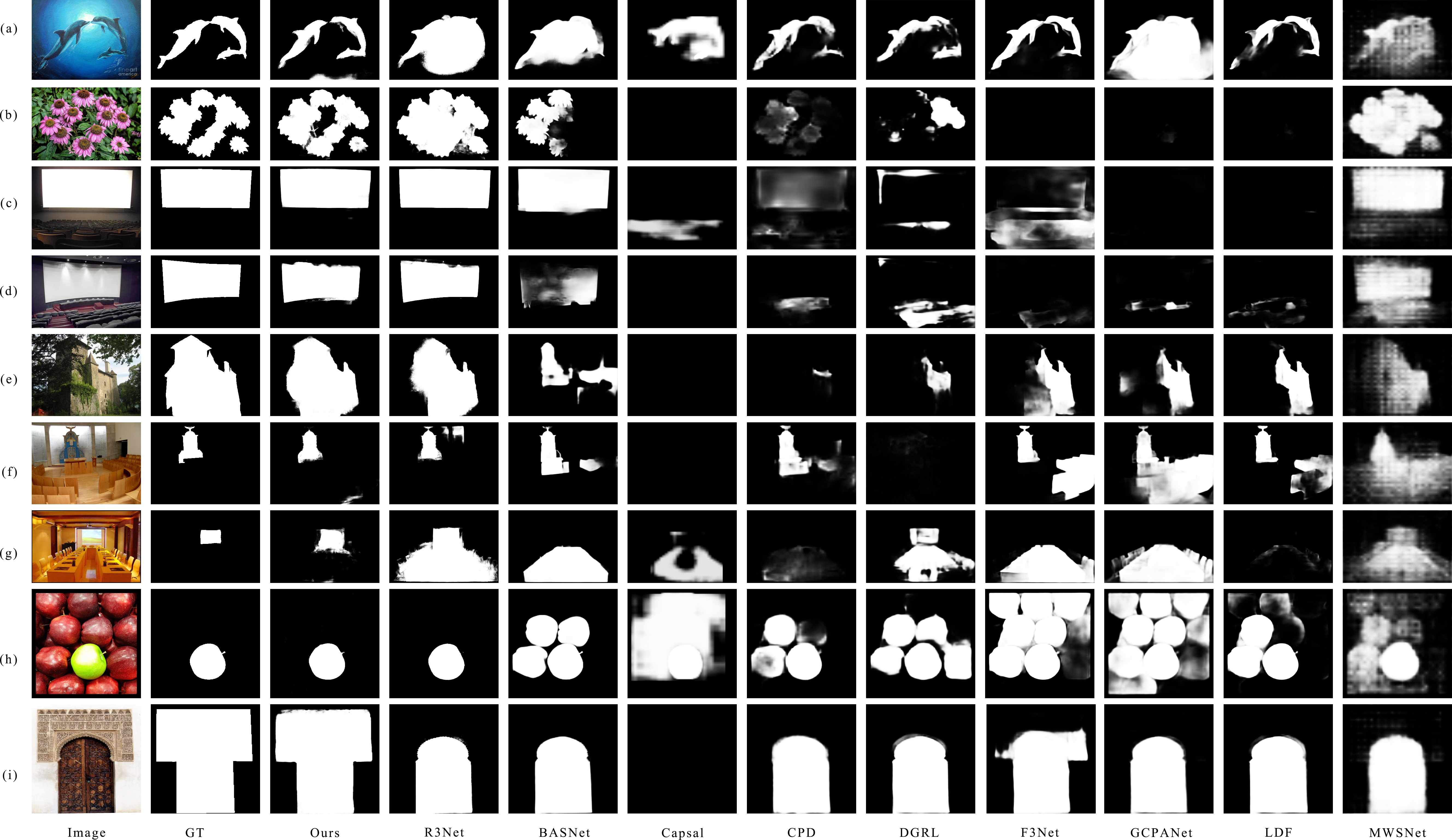}}
	\caption{Vision comparisons of 9 different methods on 8 kinds of scenarios.}
	\label{Fig7}
\end{figure*}

The comparison results of various test scenarios are shown in Figure \ref{Fig7}, including multiple objects, small objects, large objects, complex objects and cluttered background.

For multiple objects (first two rows of Figure \ref{Fig7}), our model can detect all salient objects with complete segmentation and neat edges. DGRL, CPD and Capsal are unable to obtain complete segmentation results, and MSWSNet detection results is fuzzy. R3Net and GCPANet are easily disturbed by background information, and mistakenly regard background region as salient object detection result. 

Third to fifth rows of Figure \ref{Fig7} are the comparison of large objects, the size of which is almost half of the whole image. Our model can accurately segment the edge of objects while ensuring the integrity of internal details. But other networks obtain incomplete detection results for large objects, lacking internal details.

Our method gets precise detection results in detecting small objects with complex background scenario (sixth and seventh rows of Figure \ref{Fig7}). While other models, such as LDF and BASNet, are unable to detect these small objects accurately, and contains part of background region in the segmentation results.

In the similar appearance scenario (eighth row of Figure \ref{Fig7}), there are some disturbances very similar to the features of the object in the background. Our model can detect the most distinctive object, while other approaches are difficult to eliminate the confusing disturbance of objects in the image.

Our method performs well in detecting multiple large objects (last row of Figure \ref{Fig7}). The scenario includes combined objects. Our method obtains precise and complete detection results. Most of other networks detect only one object. F3Net detects two objects, but the segmentation results are incomplete.

\subsection{Ablation Analysis}

In this paper, ablation study covers two parts, the training process with different strategy and saliency feature extractor with different orders separately.

\subsubsection{Different training strategies}

Our structure of feature extraction includes the stages of GFE and FP. The two stages are respectively responsible for extracting complete and general basic features, and modifying those features to suit for saliency tasks well. Therefore, our network no longer finetune the backbone in order to preserve the powerful ability of extracting basic features. We contrast two training strategies to demonstrate the effectiveness of our design.

We both use the ResNext101 pretrained on ImageNet as backbone in two strategies. The difference is that our method freezes the parameters of the backbone, while the contrast method does not freeze the parameters. The validation experiments are performed on five datasets.

\begin{itemize} 
	\item Quantitative comparisons
\end{itemize}
Qualitative comparisons of the MAE, MaxFm, MeanFm, Em, Sm results based on five datasets is shown in Table \ref{Table2}.
Contrasting the two training strategies on the five datasets, we find that freezing parameters during training makes the performance of MAE improve 5\% and the remaining metrics improve 10\% averagely. It illustrates the effectiveness of our training method.

\begin{figure}
	\centerline{\includegraphics[width=0.95\columnwidth]{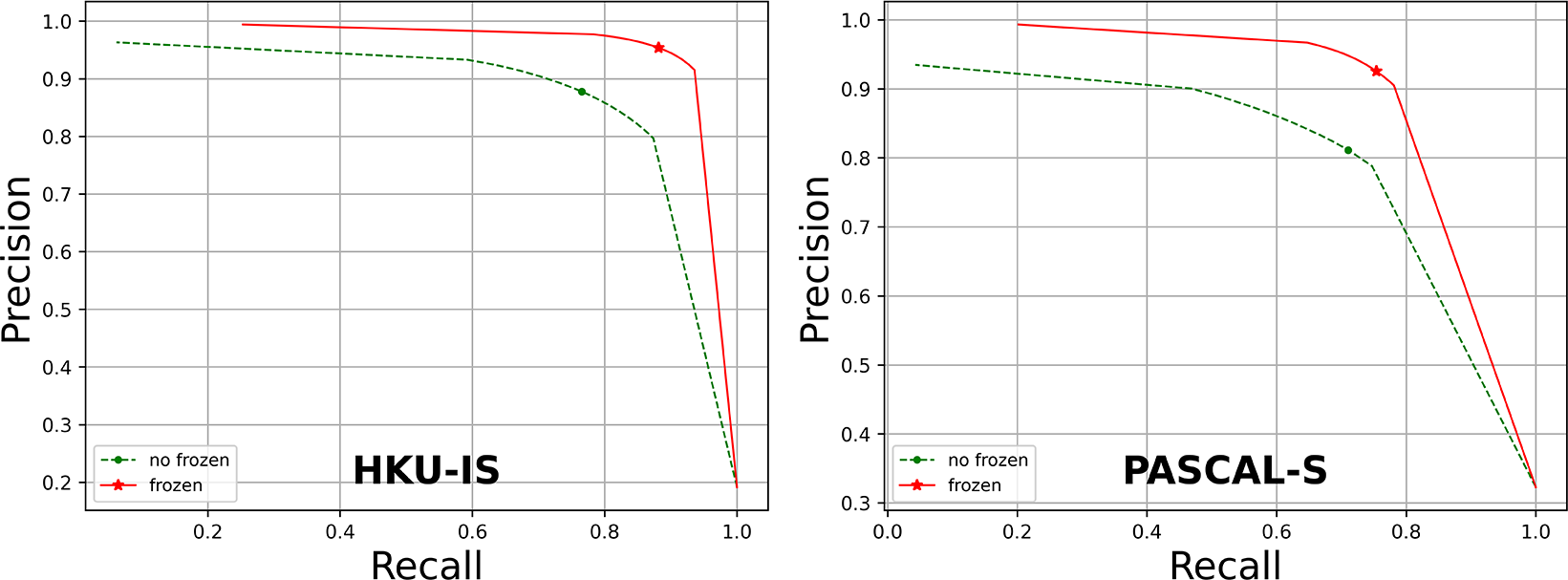}}
	\caption{Comparison of precision-recall curves on two different datasets.}
	\label{Fig8}
\end{figure}

\begin{table*}[]
	\caption{Quantitative evaluation. Comparison of different training methods. MAE (↓), MaxFm (↑), MeanFm (↑), MaxEm (↑), MeanEm (↑),SM(↑). The best result are highlighted in red.}
	\centering
	\begin{tabular}{ccccccc}
		\hline
		Datasets                    & Criteria  & MAE                                    & MaxFm                                  & MeanFm                                 & Em                                     & SM                                     \\ \hline
		& No Frozen & {\color[HTML]{FF0000} \textbf{0.0916}} & 0.6847                                 & 0.6742                                 & 0.8082                                 & 0.7546                                 \\ \cline{2-7} 
		\multirow{-2}{*}{DUT-OMRON} & Frozen    & {\color[HTML]{FF0000} \textbf{0.0916}} & {\color[HTML]{FF0000} \textbf{0.7599}} & {\color[HTML]{FF0000} \textbf{0.7543}} & {\color[HTML]{FF0000} \textbf{0.8083}} & {\color[HTML]{FF0000} \textbf{0.8246}} \\ \hline
		& No Frozen & 0.0837                                 & 0.8293                                 & 0.8206                                 & 0.8769                                 & 0.819                                  \\ \cline{2-7} 
		\multirow{-2}{*}{ECSSD}     & Frozen    & {\color[HTML]{FF0000} \textbf{0.0397}} & {\color[HTML]{FF0000} \textbf{0.9276}} & {\color[HTML]{FF0000} \textbf{0.9220}} & {\color[HTML]{FF0000} \textbf{0.9496}} & {\color[HTML]{FF0000} \textbf{0.9172}} \\ \hline
		& No Frozen & 0.1748                                 & 0.7478                                 & 0.7303                                 & 0.7759                                 & 0.6981                                 \\ \cline{2-7} 
		\multirow{-2}{*}{PASCAL-S}  & Frozen    & {\color[HTML]{FF0000} \textbf{0.1177}} & {\color[HTML]{FF0000} \textbf{0.8397}} & {\color[HTML]{FF0000} \textbf{0.8331}} & {\color[HTML]{FF0000} \textbf{0.8481}} & {\color[HTML]{FF0000} \textbf{0.7950}} \\ \hline
		& No Frozen & 0.0694                                 & 0.8244                                 & 0.8158                                 & 0.8859                                 & 0.8162                                 \\ \cline{2-7} 
		\multirow{-2}{*}{HKU-IS}    & Frozen    & {\color[HTML]{FF0000} \textbf{0.0295}} & {\color[HTML]{FF0000} \textbf{0.9283}} & {\color[HTML]{FF0000} \textbf{0.9223}} & {\color[HTML]{FF0000} \textbf{0.9610}} & {\color[HTML]{FF0000} \textbf{0.9112}} \\ \hline
		& No Frozen & 0.1718                                 & 0.6896                                 & 0.6653                                 & 0.76                                   & 0.6605                                 \\ \cline{2-7} 
		\multirow{-2}{*}{SOD}       & Frozen    & {\color[HTML]{FF0000} \textbf{0.1052}} & {\color[HTML]{FF0000} \textbf{0.8359}} & {\color[HTML]{FF0000} \textbf{0.7975}} & {\color[HTML]{FF0000} \textbf{0.8502}} & {\color[HTML]{FF0000} \textbf{0.7694}} \\ \hline
	\end{tabular}
	\label{Table2}
\end{table*}
\begin{itemize}
	\item Qualitative comparisons
\end{itemize}
We also compare the PR curves of the two training strategies on HKU-IS and PASCAL-S. As shown in Figure \ref{Fig8}, the red solid line represents the performance of freezing parameters, and the green dotted line represents the performance of non-freezing parameters. The red curve gets higher precision than green curve when in the same recall, and the length of the former is shorter. It indicates that the detection performance is much better in parameters-freezing training method than non-freezing.	
	
\begin{figure}
	\centerline{\includegraphics[width=0.95\columnwidth]{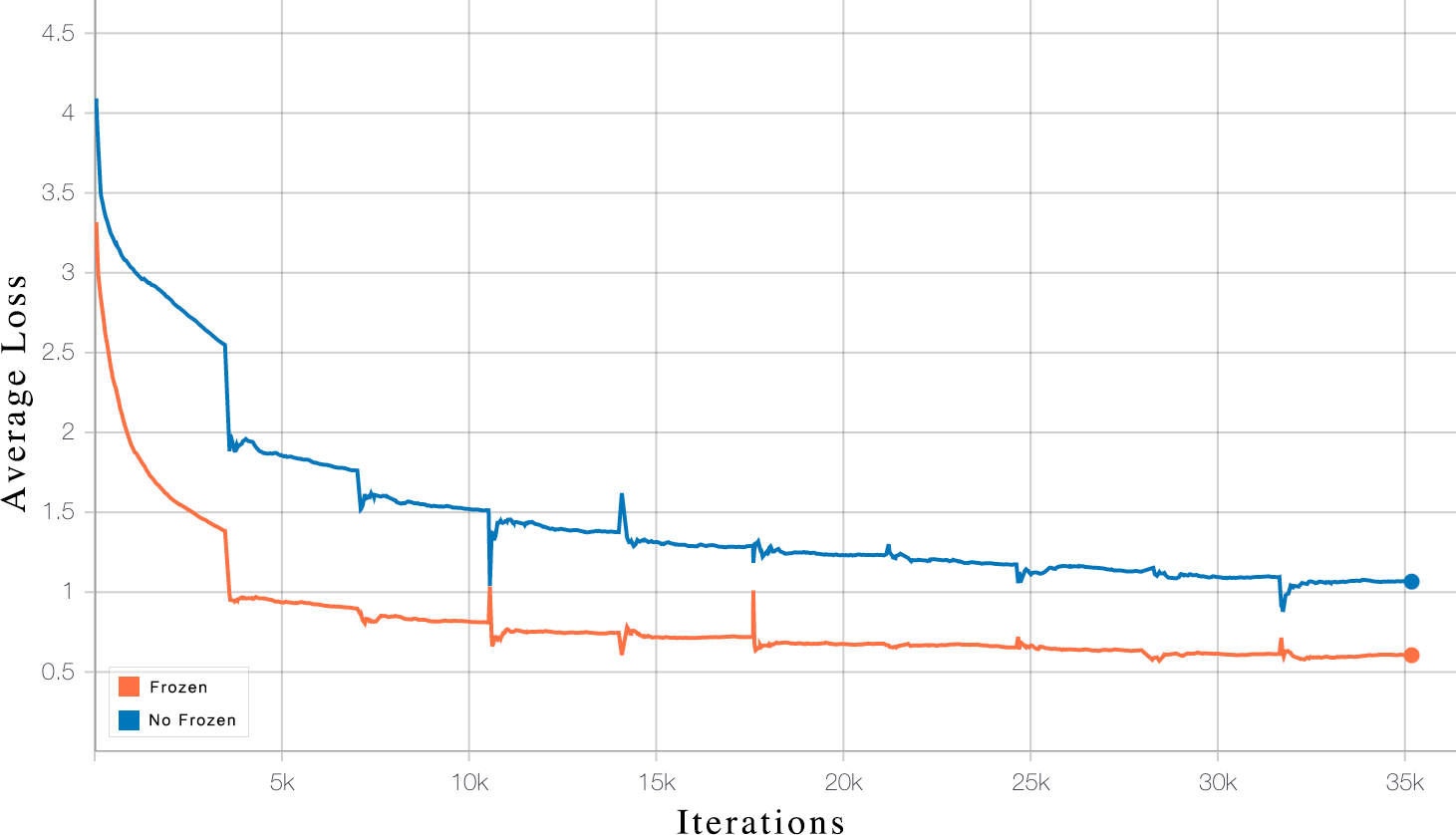}}
	\caption{Comparison of average training loss from different training methods.}
	\label{Fig9}
\end{figure}
	
We present the loss curves of the two training methods with 10 epochs in Figure \ref{Fig9}. The orange curve represents the parameters-freezing approach, while the blue curve represents the contrast approach. The orange loss curve converges faster than the blue one. Besides, the former achieves a much lower converged loss value. It shows that the training speed and the accuracy have increased a lot when freezing parameters.

\begin{figure}
	\centerline{\includegraphics[width=0.95\columnwidth]{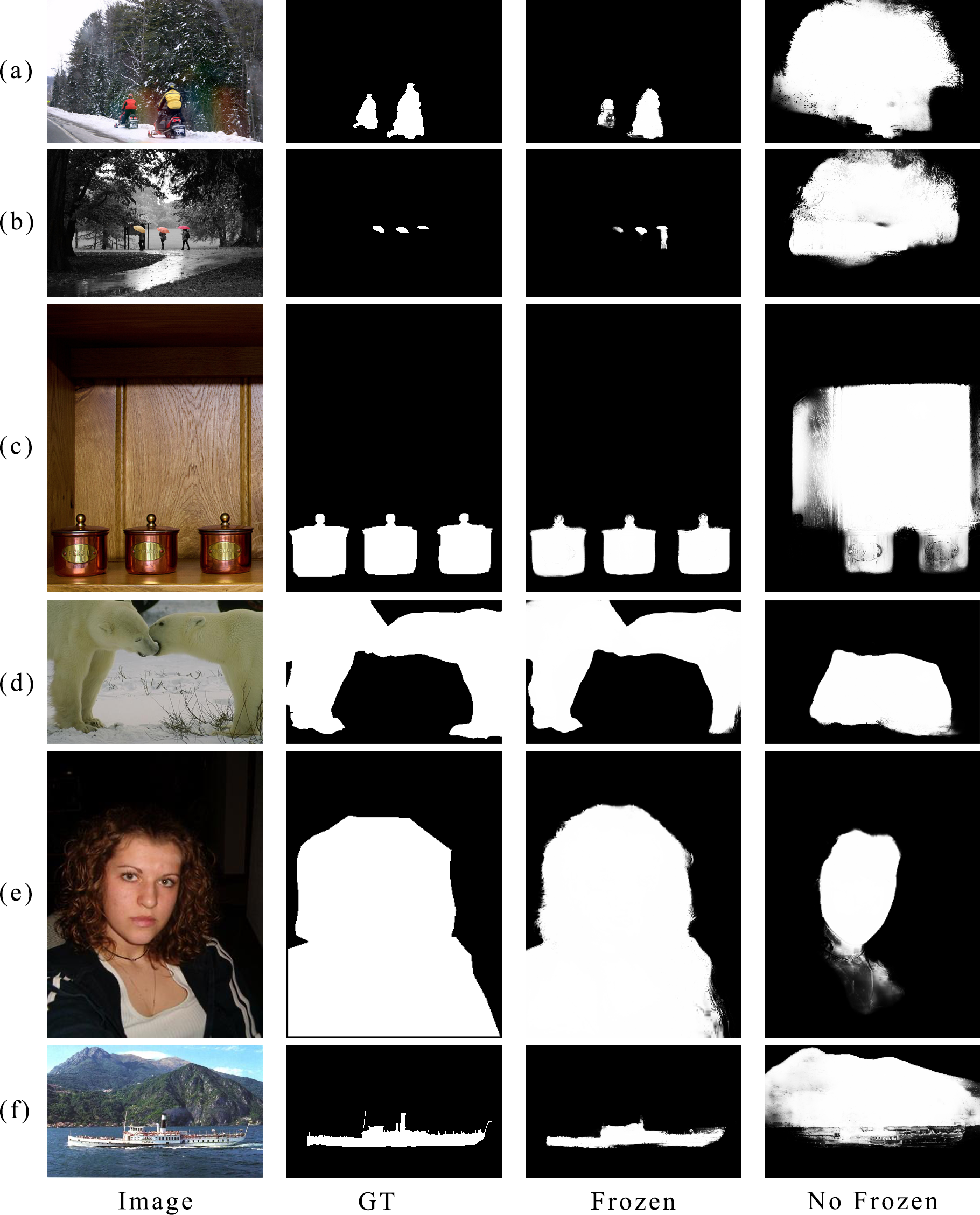}}
	\caption{Vision comparisons of two different training strategies with 6 scenarios.}
	\label{Fig10}
\end{figure}

\begin{itemize}
	\item Vision conparision
\end{itemize}
Figure \ref{Fig10} shows the vision comparison results of the two training approaches. The images include three scenarios with multiple objects, large objects and complex background. The result of single objects scenario is omitted due to good performance in the both methods.

In complex background scenario (first row of Figure \ref{Fig10}), the parameters-freezing approach can accurately segment the foreground and complex background, while the contrast method detects no foreground region at all.
For multiple objects (second to forth rows of Figure \ref{Fig10}), the approach of freezing parameters completely detects all the salient objects with accurate object boundaries, while the approach of non-freezing parameters is greatly disturbed by the background. 

The last two rows of Figure \ref{Fig10} are for large objects scenario. As we can see, the parameters-freezing method gets complete object contour, while the contrast method gets only part of the contour. 
	
\subsubsection{Different order of SFE}

To further explore the performance of the proposed network and the potential of high-order SFEMs, we design the other set of ablation experiments.

We compare three saliency feature extractors with different order SFEMs shown in Figure \ref{Fig3}. The experimental results on five test datasets are shown in Table \ref{Table3}.

\begin{itemize} 
	\item Qualitative comparisons
\end{itemize}
\begin{table*}[]
		\caption{Quantitative evaluation. Comparison of different-order SFEs. F-measure MAE (↓), MaxFm (↑), MeanFm (↑), MaxEm (↑), MeanEm (↑),SM(↑).The best three results are highlighted in red, blue and green.}
	\begin{tabular}{llccc}
		\hline
		Datasets                    & Criteria & single- order                 & two-order                     & three-order                   \\ \hline
		& MAE      & {\color[HTML]{0070C0} 0.0916} & {\color[HTML]{00B050} 0.0601} & {\color[HTML]{FF0000} 0.0541} \\ \cline{2-5} 
		& MaxFm    & {\color[HTML]{0070C0} 0.6849} & {\color[HTML]{FF0000} 0.7687} & {\color[HTML]{00B050} 0.7677} \\ \cline{2-5} 
		& MeanFm   & {\color[HTML]{0070C0} 0.6743} & {\color[HTML]{00B050} 0.7539} & {\color[HTML]{FF0000} 0.7673} \\ \cline{2-5} 
		& MaxEm    & {\color[HTML]{0070C0} 0.8083} & {\color[HTML]{00B050} 0.8676} & {\color[HTML]{FF0000} 0.8722} \\ \cline{2-5} 
		& MeanEm   & {\color[HTML]{0070C0} 0.8002} & {\color[HTML]{00B050} 0.861}  & {\color[HTML]{FF0000} 0.8677} \\ \cline{2-5} 
		\multirow{-6}{*}{DUT-OMRON} & Sm       & {\color[HTML]{0070C0} 0.7546} & {\color[HTML]{FF0000} 0.8206} & {\color[HTML]{FF0000} 0.8206} \\ \hline
		& MAE      & {\color[HTML]{0070C0} 0.0397} & {\color[HTML]{00B050} 0.0385} & {\color[HTML]{FF0000} 0.0346} \\ \cline{2-5} 
		& MaxFm    & {\color[HTML]{FF0000} 0.9276} & {\color[HTML]{0070C0} 0.9251} & {\color[HTML]{00B050} 0.9264} \\ \cline{2-5} 
		& MeanFm   & {\color[HTML]{FF0000} 0.922}  & {\color[HTML]{0070C0} 0.9172} & {\color[HTML]{00B050} 0.9198} \\ \cline{2-5} 
		& MaxEm    & {\color[HTML]{FF0000} 0.9496} & {\color[HTML]{0070C0} 0.9484} & {\color[HTML]{00B050} 0.9494} \\ \cline{2-5} 
		& MeanEm   & {\color[HTML]{0070C0} 0.9368} & {\color[HTML]{00B050} 0.9407} & {\color[HTML]{FF0000} 0.9452} \\ \cline{2-5} 
		\multirow{-6}{*}{ECSSD}     & Sm       & {\color[HTML]{0070C0} 0.9072} & {\color[HTML]{00B050} 0.9104} & {\color[HTML]{FF0000} 0.9249} \\ \hline
		& MAE      & {\color[HTML]{0070C0} 0.1177} & {\color[HTML]{00B050} 0.1124} & {\color[HTML]{FF0000} 0.1069} \\ \cline{2-5} 
		& MaxFm    & {\color[HTML]{00B050} 0.8397} & {\color[HTML]{0070C0} 0.837}  & {\color[HTML]{FF0000} 0.8473} \\ \cline{2-5} 
		& MeanFm   & {\color[HTML]{00B050} 0.8331} & {\color[HTML]{0070C0} 0.8316} & {\color[HTML]{FF0000} 0.843}  \\ \cline{2-5} 
		& MaxEm    & {\color[HTML]{0070C0} 0.8481} & {\color[HTML]{00B050} 0.8556} & {\color[HTML]{FF0000} 0.8605} \\ \cline{2-5} 
		& MeanEm   & {\color[HTML]{0070C0} 0.8228} & {\color[HTML]{00B050} 0.8318} & {\color[HTML]{FF0000} 0.8427} \\ \cline{2-5} 
		\multirow{-6}{*}{PASCAL-S}  & Sm       & {\color[HTML]{0070C0} 0.785}  & {\color[HTML]{00B050} 0.7916} & {\color[HTML]{FF0000} 0.8}    \\ \hline
		& MAE      & {\color[HTML]{0070C0} 0.0295} & {\color[HTML]{00B050} 0.0284} & {\color[HTML]{FF0000} 0.0275} \\ \cline{2-5} 
		& MaxFm    & {\color[HTML]{FF0000} 0.9283} & {\color[HTML]{0070C0} 0.9266} & {\color[HTML]{FF0000} 0.9283} \\ \cline{2-5} 
		& MeanFm   & {\color[HTML]{FF0000} 0.9223} & {\color[HTML]{0070C0} 0.917}  & {\color[HTML]{00B050} 0.92}   \\ \cline{2-5} 
		& MaxEm    & {\color[HTML]{FF0000} 0.961}  & {\color[HTML]{0070C0} 0.9597} & {\color[HTML]{00B050} 0.9599} \\ \cline{2-5} 
		& MeanEm   & {\color[HTML]{0070C0} 0.9483} & {\color[HTML]{00B050} 0.9524} & {\color[HTML]{FF0000} 0.9533} \\ \cline{2-5} 
		\multirow{-6}{*}{HKU-IS}    & Sm       & {\color[HTML]{0070C0} 0.9112} & {\color[HTML]{00B050} 0.9154} & {\color[HTML]{FF0000} 0.9172} \\ \hline
		& MAE      & {\color[HTML]{0070C0} 0.1052} & {\color[HTML]{00B050} 0.1035} & {\color[HTML]{FF0000} 0.0953} \\ \cline{2-5} 
		& MaxFm    & {\color[HTML]{0070C0} 0.8359} & {\color[HTML]{FF0000} 0.8396} & {\color[HTML]{00B050} 0.8372} \\ \cline{2-5} 
		& MeanFm   & {\color[HTML]{0070C0} 0.7975} & {\color[HTML]{00B050} 0.8031} & {\color[HTML]{FF0000} 0.8195} \\ \cline{2-5} 
		& MaxEm    & {\color[HTML]{0070C0} 0.8402} & {\color[HTML]{FF0000} 0.8596} & {\color[HTML]{00B050} 0.8593} \\ \cline{2-5} 
		& MeanEm   & {\color[HTML]{0070C0} 0.791}  & {\color[HTML]{00B050} 0.8057} & {\color[HTML]{FF0000} 0.8237} \\ \cline{2-5} 
		\multirow{-6}{*}{SOD}       & Sm       & {\color[HTML]{0070C0} 0.7594} & {\color[HTML]{00B050} 0.7708} & {\color[HTML]{FF0000} 0.7845} \\ \hline
	\end{tabular}
	\label{Table3}
\end{table*}

The performance of the network fluctuates with the order of the module increasing. The best results are obtained in three-order modules from the overall measurements. The extractor of three-order modules achieves the most significant improvement, averagely up to 8\%, on DUT-OMRON, compared with the exactor of single-order module. The images of DUT-OMRON are of low contrast between objects and background. It indicates that the network can extract deep saliency features in the image by the cascade of the modules. The performance of all metrics increases between 1\% - 2\% on PASCAL-S. And on other datasets, half of the metrics get improvements, while the remaining metrics have a few fluctuations. The results mean that our module shows good stability. The reduction of MAE on the five datasets indicates that the error of the network will gradually decreases as the order of the modules increases. The experiment proves that our model has good effect on the saliency feature extraction. We do not use 4 and other higher-order SFEMs, for the reason that we find the space for performance improvement decreases gradually as the order of the modules increases, but the parameters of the network and the size of the model increase rapidly.
\begin{itemize}
	\item Vision comparison
\end{itemize}

\begin{figure}
	\centerline{\includegraphics[width=0.95\columnwidth]{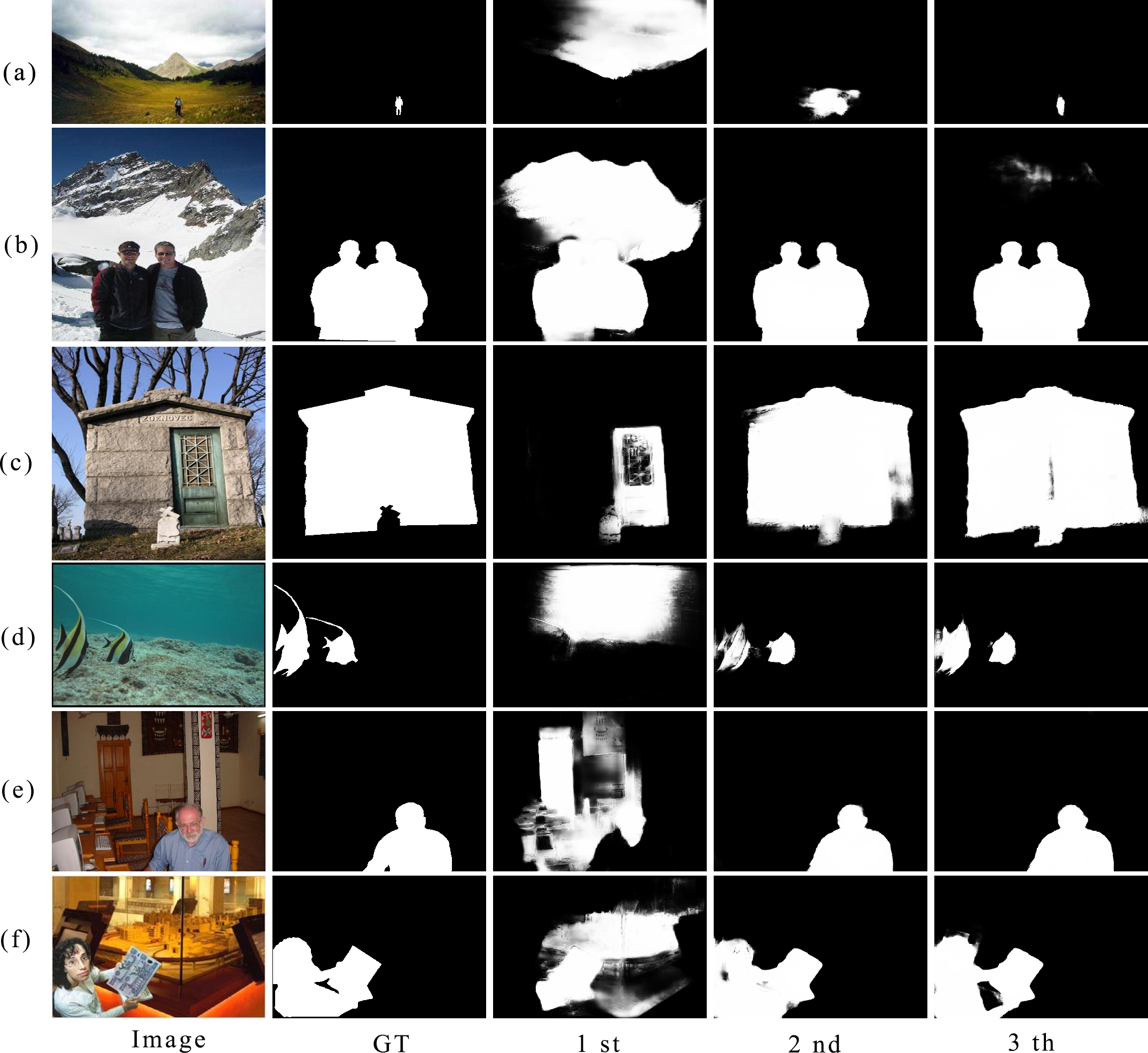}}
	\caption{Vision comparisons of different order of SFEM with 6 scenarios.}
	\label{Fig11}
\end{figure}

The comparison of experimental results is shown in Figure \ref{Fig11}. The scenarios include small objects, large objects, multiple objects, and complex single object. The results show that with single-order module, the detection is easily disturbed by the background and get incomplete segmentation for salient objects. As the order of the module increases, the network can gradually extract deep saliency features, getting more accurate segmentation maps.

For small objects (first row of Figure \ref{Fig11}). With single-order modules, the object is totally lost. With two-order modules, the network can find the approximate position of object but unable to accurately segment the region. With three-order modules, the object region can be segmented precisely.

For large objects (second row of Figure \ref{Fig11}). The detection results only contain part of the internal information of the object when using single-order modules. The object contour of segmentation is roughly complete when using two-order modules. The best detection results are obtained with three-order modules.
The third and fourth rows of Figure \ref{Fig11} show the results of multiple objects. The single-order modules cannot segment multiple objects completely, and mistakenly detect part of background region. The better results are obtained by higher-order modules.

In this scene of complex background (last two rows of Figure \ref{Fig11}), the results almost contain only the background when using single-order modules. The segmentation of the object contour and internal details are relatively accurate when using higher-order modules.

\section{Conclusion}

We propose a four-stage framework for salient object detection using High-order contrast operator (PAANet) applied in natural scenaries. Our method draws on the visual perception mechanism. The first two stages of the network can separately extract general basic features and process them to suit for SOD tasks. The backbone uses a model pre-trained in ImageNet, which needs not to finetune. At the same time, the SFEMs in this article can fully extract the contrast features in the image through the efficient cascading design, which also alleviate the problem of the uncertainty of saliency. The final results are obtained through multi-scale feature aggregation. Experiments have indicated that the PAANet outperforms the current best methods in several datasets.

In future work, we will extend our framework and training strategy to other fields of visual tasks not limited to SOD, based on HVS. We attempt to make it become a general framework for object detection based on visual perception mechanism. According to different requirements and confinements of vision tasks, the relative stages of the network can be adjusted adaptively, using the process of \textit{general basic features extraction} - \textit{feature preprocessing} - \textit{special task features extraction} - \textit{decision} to realize the corresponding object detection task.

\bibliography{MyCollection}

\end{document}